%% file: main.tex
\documentclass[12pt]{interact}

\usepackage{epstopdf}
\usepackage[caption=false]{subfig}

\usepackage{diagbox}
\usepackage{multirow}
\usepackage{color}
\usepackage{url}

\newif\ifarxiv
\arxivtrue

\usepackage[longnamesfirst,sort]{natbib}
\bibpunct[, ]{(}{)}{;}{a}{,}{,}
\usepackage{lineno}
\linenumbers

\renewcommand\bibfont{\fontsize{10}{12}\selectfont}

\newcommand{\figcaption}[1]{\def\@captype{figure}\caption{#1}}
\newcommand{\tblcaption}[1]{\def\@captype{table}\caption{#1}}

\theoremstyle{plain}

\theoremstyle{definition}

\theoremstyle{remark}

\usepackage{lineno}
\newcommand*\patchAmsMathEnvironmentForLineno[1]{
  \expandafter\let\csname old#1\expandafter\endcsname\csname #1\endcsname
  \expandafter\let\csname oldend#1\expandafter\endcsname\csname end#1\endcsname
  \renewenvironment{#1}
     {\linenomath\csname old#1\endcsname}
     {\csname oldend#1\endcsname\endlinenomath}}
\newcommand*\patchBothAmsMathEnvironmentsForLineno[1]{
  \patchAmsMathEnvironmentForLineno{#1}
  \patchAmsMathEnvironmentForLineno{#1*}}
\AtBeginDocument{
\patchBothAmsMathEnvironmentsForLineno{equation}
\patchBothAmsMathEnvironmentsForLineno{align}
\patchBothAmsMathEnvironmentsForLineno{flalign}
\patchBothAmsMathEnvironmentsForLineno{alignat}
\patchBothAmsMathEnvironmentsForLineno{gather}
\patchBothAmsMathEnvironmentsForLineno{multline}
}
\ifarxiv
\nolinenumbers
\else
\linenumbers
\fi

\begin{document}

\articletype{ORIGINAL ARTICLE}

\title{Location analysis of players in UEFA EURO 2020 and 2022 using generalized valuation of defense by estimating probabilities}

\ifarxiv
\author{
\name{Rikuhei Umemoto\textsuperscript{a}, Kazushi Tsutsui\textsuperscript{a}, and Keisuke Fujii\textsuperscript{a,b,c}\thanks{corresponding author:  fujii@i.nagoya-u.ac.jp}}
\affil{\textsuperscript{a}Graduate School of Informatics, Nagoya University, Nagoya, Aichi, Japan; \textsuperscript{b}RIKEN Center for Advanced Intelligence Project, Fukuoka, Fukuoka, Japan; \textsuperscript{c}PRESTO, Japan Science and Technology Agency, Kawaguchi, Saitama, Japan}
}
\else
\author{Anonymous}
\fi
\maketitle

\begin{abstract}
Analyzing defenses in team sports is generally challenging because of the limited event data. 
Researchers have previously proposed methods to evaluate football team defense by predicting the events of ball gain and being attacked using locations of all players and the ball.
However, they did not consider the importance of the events, assumed the perfect observation of all 22 players, and did not fully investigated the influence of the diversity (e.g., nationality and sex). 
Here, we propose a generalized valuation method of defensive teams by score-scaling the predicted probabilities of the events.
Using the open-source location data of all players in broadcast video frames in football games of men’s Euro 2020 and women’s Euro 2022, we investigated the effect of the number of players on the prediction and validated our approach by analyzing the games. 
Results show that for the predictions of being attacked, scoring, and conceding, all players’ information was not necessary, while that of ball gain required information on three to four offensive and defensive players.
With game analyses we explained the excellence in defense of finalist teams in Euro 2020. 
Our approach might be applicable to location data from broadcast video frames in football games. 

\end{abstract}


\section{Introduction}
Soccer or association football has one of the highest number of players and spectators in the world.
Because of the diversity, we need to establish a common indicator to quantitatively evaluate soccer teams, regardless of the background of the constituent players.
However, analysis of soccer data is usually difficult because the ball and players are moving continuously during a game and specific event data (e.g., scores and concedes) are likely to be limited.
In particular, defensive tactics are considered difficult to evaluate because of the limited amount of available statistics, such as goals scored in the case of attacks (for overview, see a review of \cite{fujii2021data}).

Although publishing of tracking technologies and videos \citep{scott2022soccertrack, Cioppa_2022_CVPR} have recently progressed, a large amount of accurate all-time tracking data for all players remain unpublished. 
Therefore, many evaluations of attacking players with the ball are performed from event data and the coordinates of the ball. 
For example, based on scoring prediction using ball location data, researchers evaluated plays such as using the expected values of goals scored and conceded (\citet{rudd2011framework, Decroos19,Liu2020}, and others are reviewed in \citet{vanroy2020valuing}).
Although there have been some studies on off-ball player evaluation  \citep{spearman2018beyond,Fernandez18,teranishi2022evaluation} which require positional data for all players, these are usually private (i.e., not published). 
Recently, the location of all players in broadcast video frames of every event have been published by StatsBomb Inc. (a UK football analytics and data visualization company) in soccer games of men’s Euro 2020 and women’s Euro 2022 competitions. 
For example, using this data including a pass receiver and defenders, researchers developed a machine learning method for prediction and valuation of penetrative passes \citep{rahimian2022lets}.

However, there have been a few studies on defensive play evaluation.
For example, using only ball location and event data, researchers have evaluated interceptions \citep{Piersma20} and the effectiveness of defensive plays by the expected value of a goal-scoring opportunity conceded \citep{Robberechts19}.
For team evaluation using all players' tracking data, \citet{toda2022evaluation} have previously proposed a method to evaluate team defense by predicting ball gain (or recovery) and being attacked.
However, this approach has the following limitations: (1) they did not consider the importance weight of the ball gain and being attacked, (2) they assumed the perfect observation of all 22 players and the effect of the lack of the observation is unknown, and (3) they only investigated in a domestic male professional league and did not fully investigated in terms of the diversity (e.g., nations and sexes). 

The purpose of this study is to address these issues. 
First, we propose a generalized valuation method of defensive teams by score-scaling the predicted probabilities of the events called \textit{Generalized Valuing Defense by Estimating Probabilities} (GVDEP). 
Then, using the location of all players in the broadcast video frames of every event in football games of men’s Euro 2020 and women’s Euro 2022, we investigated the effect on the prediction models of reducing the number of players and validated our approach to analyze the games of each team. 
The main contributions of this work are as follows: (i) We generalize the previous team defense valuation method \citep{toda2022evaluation} by weighting the probabilities of ball gain and being attacked based on the prediction of scoring and conceding \citep{Decroos19}. (ii) We verified the classifiers for prediction of ball gain, being attacked, scoring, and conceding when we reduced the number of players. (iii) We verified our method using a diverse soccer dataset in terms of  nations and both sexes (open data of men’s Euro 2020 and women’s Euro 2022). 
Our approach can encourage even non-professionals watching broadcast videos to discuss the valuation of defensive teams at a particular event.
Furthermore, our approach can provide quantitative and useful indicator for valuing and scouting our and opponent teams in defense.

\section*{Materials and methods}
\subsection*{Dataset}
Since having the location data of players of teams in diverse nations, we chose the open-source data with all games of UEFA men's Euro 2020 (51 games) and women's Euro 2022 (31 games) provided from StatsBomb Inc. (UK). 
This datasets include event data (i.e., labels of actions, such as passing and shooting, and the simultaneous xy coordinates of the ball) and location data (i.e., xy coordinates) of all players ``in the frame of broadcast video'' around every event.
Many scenes of a soccer broadcast video do not show all 22 players; therefore, as shown in Figure \ref{fig:example}, note that some of datasets do not include all 22 players' information.
In the case of this dataset, the min, the first quartile, the median, the third quartile and the max are 0.0, 11.0, 15.0, 18.0 and 22.0 respectively about the number of players in a scene (for the histogram, see Supplementary Figure 5).
Also, we used ``event'' as the above meaning based on the previous studies of \cite{Decroos19,pappalardo2019playerank}.
Data acquisition was based on the contract between the competitions (UEFA Euro2020/2022) and the company (StatsBomb, Inc), not between the players and us.
The Statsbomb dataset has been widely used in academic studies (e.g., \cite{gregory2022influence, Decroos19}).
It would be guaranteed that the use of the data would not infringe on any rights of players or teams.
The dataset is available at \url{https://github.com/statsbomb/open-data}.

In all 51 games of UEFA EURO 2020, 
142 of the 1262 shots were on goal, 7,027 effective attacks were played, and 2,463 ball gains were realized.
Similarly, in all 31 games of UEFA EURO 2022, 
95 of the 880 shots were on goal, 4,717 effective attacks were played, and 1,839 ball gains were realized.
An effective attack is defined as an event 
that ends in shooting or penetrating into the penalty area. 
Also, ball gains are defined as a change in the attacking by some factors such as tackle, interception and offside.
In this study, an effective attack is defined as \textit{attacked} from viewpoints of defenders.
It should be noted that we labeled for each event (not an attack segment) whether positive/negative attacked/ball gains occurred at an event.


\subsection*{Proposed Method}
The flow diagram of our method is shown in Figure \ref{fig:diagram}. 
We perform data pre-processing and feature creation, training classifiers, prediction with the classifiers, and computing GVDEP as well as Valuing Defense by Estmating Probabilities (VDEP) proposed by \cite{toda2022evaluation}.
Here we describe the core idea of our proposed method and the details are described in the following subsection.
Suppose that the state of the game is given by $S = [s_1,\ldots,s_N]$ in chronological order.
We consider $s_i$, like \cite{toda2022evaluation}, as the $i$th state.
This includes features about the event or action $a_i$, the on-ball features $b_i$ and the off-ball features $o_i$ not close to the ball at the time of the action such as attacking/defending players' coordinates at a state.
Note that we show the result about the validation of the number of the players in the following section.
To evaluate all state transition for defensive and offensive actions in this study from the defender's perspective, the following time index $i$ is used as the $i$th \textit{event}.

As with \cite{toda2022evaluation}, we define the probability of future ball gain $P_{gains}(s_i)$ and attacked $P_{attacked}(s_i)$ in the game state $s_i$ of a certain interval at an $i$th event. 
These probabilities were given by the classifier trained.
We focus on whether changing the game state affects the team defense 
like the work of \cite{Decroos19}, which directly used these probabilities for state evaluation. 
For this reason, first we use the difference between the probabilities at a state $s_i$ and ones at previous state $s_{i-1}$ as follows:
\begin{align}
\label{eq:deltas}
    \Delta P_{gains}(s_i, x) &= P_{gains}(s_i, x) - P_{gains}(s_{i-1}, x),
    \\
    \Delta P_{attacked}(s_i, x) &= P_{attacked}(s_i, x) - P_{attacked}(s_{i-1}, x),
\end{align}
where $x$ is which team is defending.

The most critical problem in the previous 
was constant parameter to balance the probabilities of ball gain and being attacked, which was defined by the frequencies of both.
To appropriately weight both variables in a score scale, the value of defense in the proposed method $V_{gvdep}$ is weighted with VAEP of \cite{Decroos19} at a time when ball gains or attacked as follows:

\begin{align}
    weight\_gains &= \sum_{j \in Ev_{gains}}\frac{sign(Teams_{j})V_{vaep}(s_j)}{|Ev_{gains}|}, \\
    weight\_attacked &= \sum_{j \in Ev_{attacked}}\frac{sign(Teams_{j})V_{vaep}(s_j)}{|Ev_{attacked}|},
\end{align}
\begin{equation}
\label{eq:vGVDEP}
    V_{gvdep}(s_i) = weight\_gains \times \Delta P_{gains}(s_i) - weight\_attacked \times \Delta P_{attacked}(s_i),
\end{equation}

where each of $Ev_{gains}$ and $Ev_{attacked}$ (also $|Ev_{gains}|$ and $|Ev_{attacked}|$) is the number of ball gains/attacked in all games.
$sign(Teams_{j})$ means the sign of which team performed ball gain or attacked at a $j$th event.
Where the first term in equation (\ref{eq:vGVDEP}) can be regarded as the difference in the probabilities of scoring or conceding after gaining the ball, and the second term as that after being attacked.
Furthermore, to use the framework of \cite{Decroos19} into our formula, we 
calculated VAEP as follows:
\begin{align}
\label{eq:VAEP}
    \Delta P_{scores}(s_i) &= P_{scores}(s_i) - P_{scores}(s_{i-1})
    \\
    \Delta P_{concedes}(s_i) &= P_{concedes}(s_i) - P_{concedes}(s_{i-1})
    \\
    V_{vaep}(s_i) &= \Delta P_{scores}(s_i) - \Delta P_{concedes}(s_i)
\end{align}

To evaluate the team defense, we define $R_{gvdep}(p)$ as the evaluation value per game for team $p$ as follows:
\begin{equation}
\label{eq:rGVDEP}
    R_{gvdep}(p) = \frac{1}{M}\Sigma_{s_i\in \bm{S}_{M}^p} V_{gvdep}(s_i),
\end{equation}
where $M$ is the number of events for team $p$ defending in all matches, and $\bm{S}_M^p$ is the set of states $S$ of team $p$ defending in all matches.

\subsection*{Feature Creation}
\begin{figure}
    \begin{minipage}{0.5\textwidth}
        \centering
        \makeatletter
        \def\@captype{table}
        \makeatother
        \begin{tabular}{cccc}
        \hline
        event & player & action & GVDEP \\
        \hline \\
        1 & Declan Rice & pass & -0.004 \\
        2 & Harry Maguire & pass & -0.006 \\
        3 & Jordan Pickford & pass & 0.017
        \end{tabular}
        \label{label 2}
    \end{minipage}
    \begin{minipage}{0.6\textwidth}
        \centering
        \includegraphics[scale=0.6]{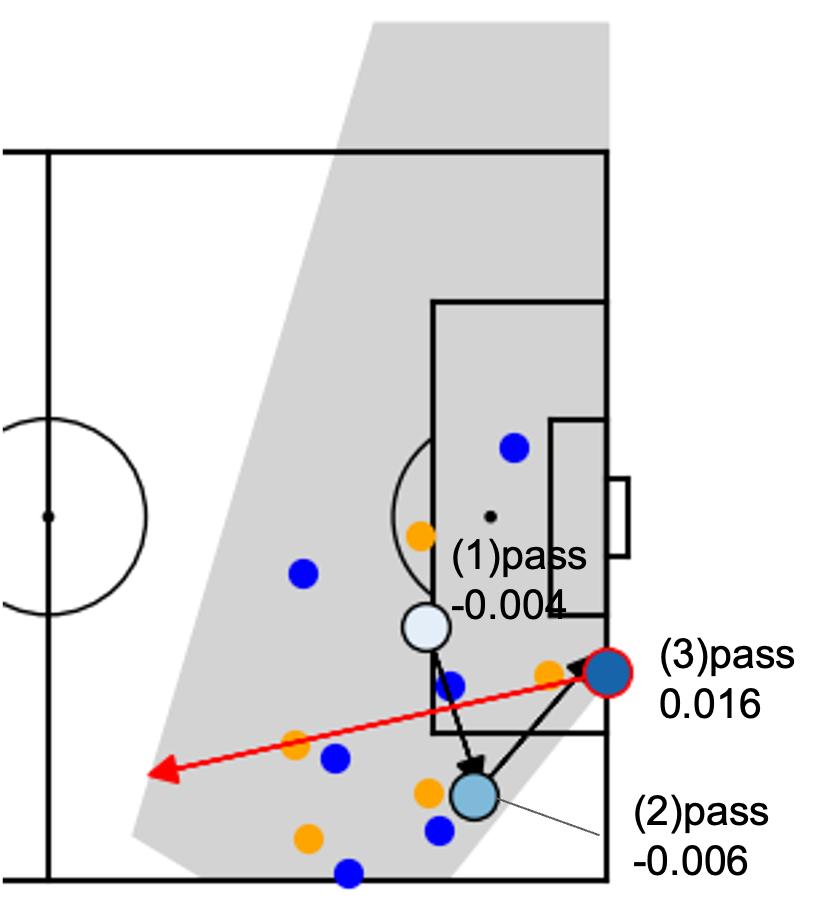}
        \label{label 1}
    \end{minipage}
    \caption{{\bf Example of a broadcast scene.}
    This figure shows the x and y coordinates of the positions of the players taken from a broadcast. Blue and orange markers, black or red arrow, and grey area mean England's and the opposite side's players, a successful or failed pass and an area projected by a broadcast video respectively. The series of scenes described that Jordan Pickford kicked the ball and reduced the probability of the opposing team penetrating into the penalty area. As a result, the value of GVDEP was high.
    }
\label{fig:example}
\end{figure}

\begin{figure}
    \centering
    \includegraphics[scale=0.6]{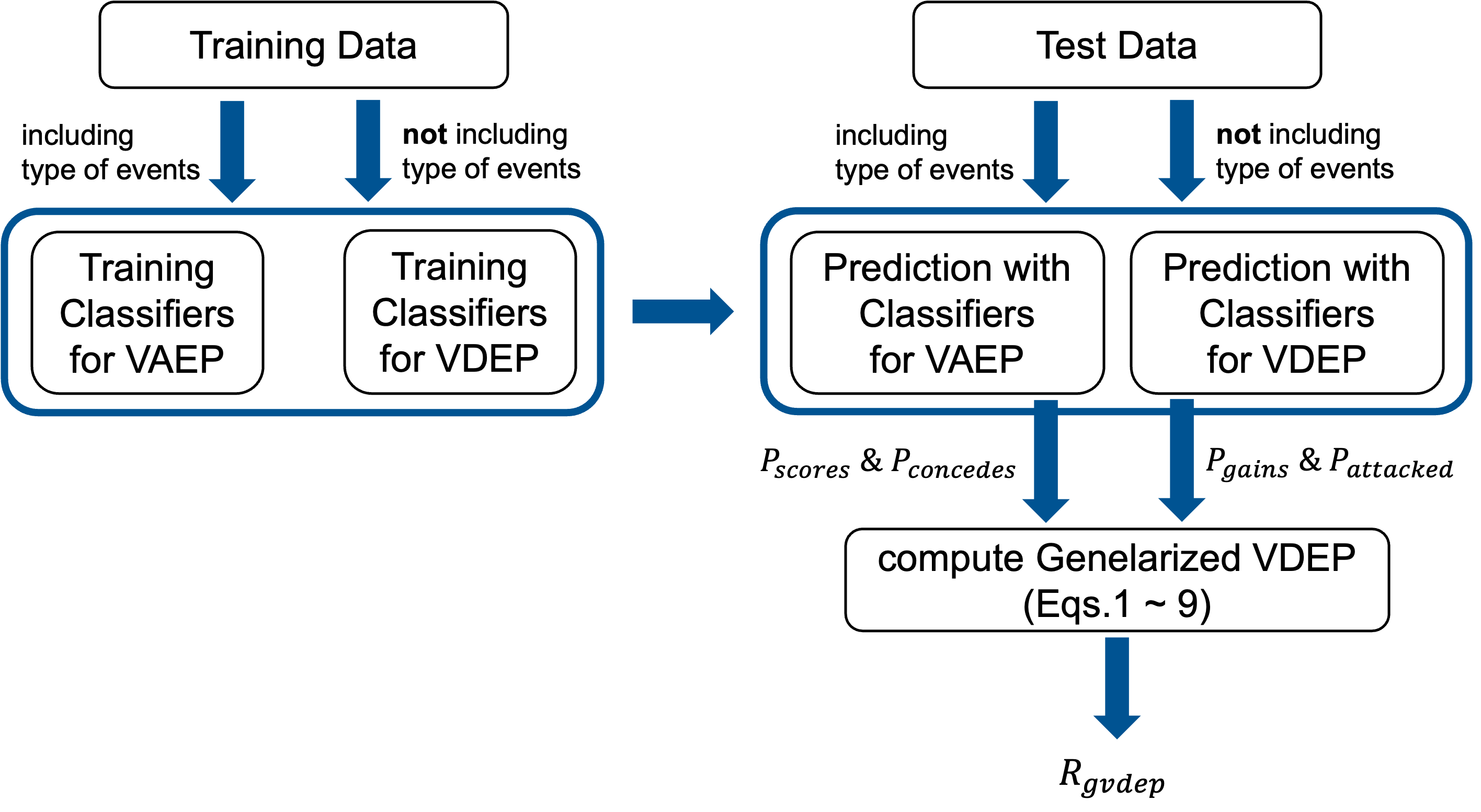}
    \caption{\label{fig:diagram}{\bf The flow diagram of the analysis.} The computation of our Generalized VDEP (GVDEP) is composed of four steps: pre-processing and feature creation, training classifiers, prediction with the classifiers, and computing VDEP.
    First, we preprocess data and create features including xy coordinates and event data for the classifier.
    Second, we train two classifiers; one is for scores and concedes, the other is for ball gains and attacked.
    The former uses data including types of events, but the latter does not include it.
    This reason is explained in the section \textit{Feature Creation}.
    Third, each trained classifier predicts scores, concedes, ball gains and attacked using xy coordinates in the test dataset. Finally, we computed GVDEP using Equations \ref{eq:deltas} to \ref{eq:rGVDEP}.
    }
\end{figure}

In this subsection, we describe pre-processing and feature creation.
As mentioned above, we used the features at a state $s = [a,b,o]$ including the $i$th and $i-1$th events for the same reasons as in the framework of \cite{toda2022evaluation}.
Using these features, 
we trained four classifiers and we predicted $P_{scores}(s_i)$ and $P_{concedes}(s_i)$, $P_{gains}(s_i)$ and $P_{attacked}(s_i)$.
Each of the classifiers estimates whether a state $s_i$ is labeled positive ($= 1$) or negative ($= 0$).
Positive label was assigned if a attacking team in a state $s_i$ scored or conceded in the subsequent $k’$ events, and the latter was labeled positive if the defending team in the state $s_i$ gained the ball or attacked in the subsequent $k$ events.
An illustrative example is shown Figure \ref{fig:example}.
In both classifications, $k$ is a parameter freely determined by the user. 
The smaller $k$ is, the shorter term the prediction is and the smaller positive labels, so we can predict the probabilities reliably and obtain unambiguous interpretation.
On the other hand, the larger $k$ is, the longer term the prediction is and the larger positive labels, so we can predict these considering many factors but obtain ambiguous interpretation. 
Since it is intrinsically difficult to solve this trade-off, we set $k=5$ and $k’=10$ like the previous studies of \cite{toda2022evaluation} and \cite{Decroos19} respectively.

Next, we describe the feature vector creation.
We first created three types of features; events, on-ball features and off-ball features.
The events' types defined by VAEP \citep{Decroos19} had 24 events: id, pass, cross, throw in, freekick crossed, freekick short, corner crossed, corner short, take on, foul, tackle, interception, shot, shot penalty, shot freekick, keeper save, keeper claim, keeper punch, keeper pick up, clearance, bad touch, non action, dribble, goalkick (for details, see the paper \citep{Decroos19}).
Yet, as shown in Figure \ref{fig:diagram}, this type was used in the classifiers for VAEP but not used for VDEP due to the difference of each concept.
Whereas we predicts future actions in VAEP, the prediction in VDEP was whether a team was able to gain the ball or was penetrated into the penalty area.
Hence, feature leakage (to know ground truth for prediction in advance) might occur if we used data including the types of events into the classifiers for VDEP.
On the other hand, we utilized on-ball features and off-ball features in the classifiers for VAEP and VDEP.
We first created the 21 dimensional features of $i$th and $i-1$th events about on-ball features.
We used the body part such as foot, head, other and head/other when the player acts at $i$th event (4 dim.), whether a yellow/red card is given at the event (2 dim.), the scoreboard and the goal difference at the event (3 dim.), the xy coordinates of the ball at the start and the end (4 dim.), the displacements of the ball from the start to the end (x, y, the Euclidean norm: 3 dim.), the distance and angle between the ball and the goal at the start and the end (4 dim.), whether a team possessing the ball is a visitor or not (1 dim.)
In addition, we considered the off-ball feature $o_i$ at the time that the event occurred in the state $s_i$.
It should be noted that data utilized in this study was provided from broadcast videos shown in Figure \ref{fig:example}.
Thus, we first verified impacts on the prediction score by changing offensive/defensive players in the order of closest to the ball.
This verification is described in the subsection \textit{Evaluation and Statistical Analysis}.
Moreover, we used the x and y coordinates of positions of all players (22 players xy coordinates)
Furthermore, we calculated the distance and the angle of each player from the ball and sorted the features in the order of closest to the ball.
Therefore, the feature for VAEP has 133 dimensions in total ($24 + 21 + 22\times 4$) and for VDEP has 109 ($21 + 22 \times 4$).



As with the previous studies by \cite{Decroos19} and \cite{toda2022evaluation},
we used $k'=10$ when calculating $P_{scores}$ and $P_{concedes}$ and $k=5$ when calculating $P_{gains}$ and $P_{attacked}$.
Also, 12,262 out of the 112,590 events generated by all teams had no data about all players' xy coordinates.
Thus, we removed its data and utilized 100,328 events for classifiers.
Out of the number of events, defined by $k$ above, there were 1,101 positive cases of scores, 186 concedes, 3,723 ball gain and 11,895 attacked.
In UEFA EURO 2022, we removed 28,218 out of 61,433 events and utilized 33,215 for the classifiers, including 454 positive cases of scores, 132 concedes, 2,551 ball gain and 5,401 attacked.
As with the previous studies, we can interpret these as goals scored and conceded are rare events compared to ball gains and attacked. 
Therefore, to verify the goals scored and conceded in this study with a smaller dataset (compared with the larger dataset in the previous work \cite{Decroos19}), we used a indicator and described it in the subsection \textit{Evaluation and Statistical Analysis}.

\subsection*{Prediction Model Implementation}
According to the previous frameworks of \cite{Decroos19} and \cite{toda2022evaluation}, we adopted XGBoost (eXtreme Gradient Boosting) proposed by \cite{Chen16} as the classifiers to predict scores, concedes, ball gains and attacked. 
Gradient tree boosting has been a popular technique since \cite{friedman2001greedy} proposed.
This technique is a prediction algorithm that uses a previous information to optimise the splitting of explanatory variables and construct a new decision tree at each stage.
Also, this performs well on a variety of areas, and in a faster and more scalable way.
Moreover, even if the prediction model itself do not consider the time series structure, the previous studies of \cite{Decroos19} proposed the way for prediction models to reflect the history of the input ($i$th and $i-1$th events) and that of the output (the subsequent $k$ events) .

In calculating VDEP and VAEP values, we verified the classifiers using a 10-fold cross-validation procedure. 
Here we define the terms of training, validation, and test (datasets). 
We train the machine learning model using the training dataset, 
validate the model performance using the validation dataset (sometimes for determining some hyper-parameters), and finally test the model performance using the test dataset.
Note that we remove data not including the x and y coordinates of positions of players from these dataset.
Such procedure benefits us to verify a model which can test the performance using a new test data (not used during training). 
In our case, the validation data was not used and hyperparameters are predetermined as default in Python library ``xgboost'' (version 1.4.1).

Our all computations were performed using Python (version 3.7.13).
In particular, the code we used 
\ifarxiv
is provided at \url{https://github.com/Rikuhei-ynwa/Generalized-VDEP}.
\else
will be shared if our manuscript is ready for publication. 
\fi

\subsection*{Evaluation and Statistical Analysis} \label{subsec:eval}
First, we verified how many attacker/defender we need to predict the above probabilities.
To this end, we defined the number of nearest attackers/defenders to the ball as \textit{n\_nearest}.
This value was $0 \leq n\_nearest \leq 11$ and by each of this value, we predicted the probabilities such as $P_{gains}$ and $P_{attacked}$.
Here we used a 10-fold cross-validation procedure.
Again, the datasets have 51 games in UEFA EURO 2020 and 31 in UEFA EURO 2022.
Hence, in the step of verification of classifiers, 9 out of 10 times we repeated the learning of classifiers using the data of 46 games and a prediction using the data of 5 games, and finally used the data of 45 games into the learning of classifiers and 6 games into the prediction (i.e., data of all 51 games were finally predicted and evaluated) to analyze all games in UEFA EURO 2020.
Similarly, in UEFA EURO 2022, first 9 times the data of 26 games was used into the learning of classifiers and 5 games into the prediction repeatedly, and last time we change 26 to 25 at the classifier stage and 5 to 6 at the prediction stage.
To validate the classifiers for the predictions, we used the F1 score as with the previous study of \cite{toda2022evaluation}.
Data in this study was much more negative than positive cases like ones used in the previous studies.
Then, the (intuitive) accuracy score may not be better when there are extremely more negative than positive cases, as in this and previous studies.
For example, the accuracy of attacked in VDEP and scored in VAEP will be $1-11,895/100,328\approx0.881$ and $1-186/100,328\approx0.998$ when all negative cases are predicted. 
For this reason, in this study, we used the F1 score to evaluate whether the true positives can be classified without considering the true negatives. 
The F1 score is the harmonic mean of Precision and Recall as follows.
\begin{equation*}
    \text{F1score} = (2 \times \text{Precision} \times \text{Recall}) / (\text{Precision} + \text{Recall})
\end{equation*}
where the Recall and the Precision is defined as the true-positive rate and the ratio of the sum of true positives and true negatives to false positives respectively.
In this index, only true positives are evaluated, not true negatives.
Therefore, we saw how changing $n\_nearest$ affected values of F1 score for $P_{scores}(s_i)$, $P_{concedes}(s_i)$, $P_{gains}(s_i)$ and $P_{attacked}(s_i)$.

For the evaluation of defense using GVDEP calculated by the above probabilities, we present examples to quantitatively and qualitatively evaluate games in a competition.
To calculate GVDEP values, the classifiers first learned with all 51 games (UEFA EURO 2020) or 31 games (UEFA EURO 2022) and tested themselves.
Note that, we performed the predictions and analyzed 36 group stage matches and 8 matches of the first round in UEFA EURO 2022 to analyze the best 16 teams with the same number of games.
Similarly, in UEFA EURO 2022, 24 group stage games and 4 quarter-final games were analyzed for the same reasons.
The GVDEP is then calculated using the probabilities obtained by the tests and equations \ref{eq:deltas} to \ref{eq:rGVDEP}.

For correlation analysis between variables, we computed Pearson's correlation coefficient $r$. 
$p<0.05$ was considered significant for all statistical analysis. 
all statistical analyses were performed using SciPy (version 1.5.4) in the Python library.

\section*{Results} \label{sec:results}
\subsection*{Verification of classifiers}

\begin{figure}
    \centering
    \includegraphics[scale=0.7]{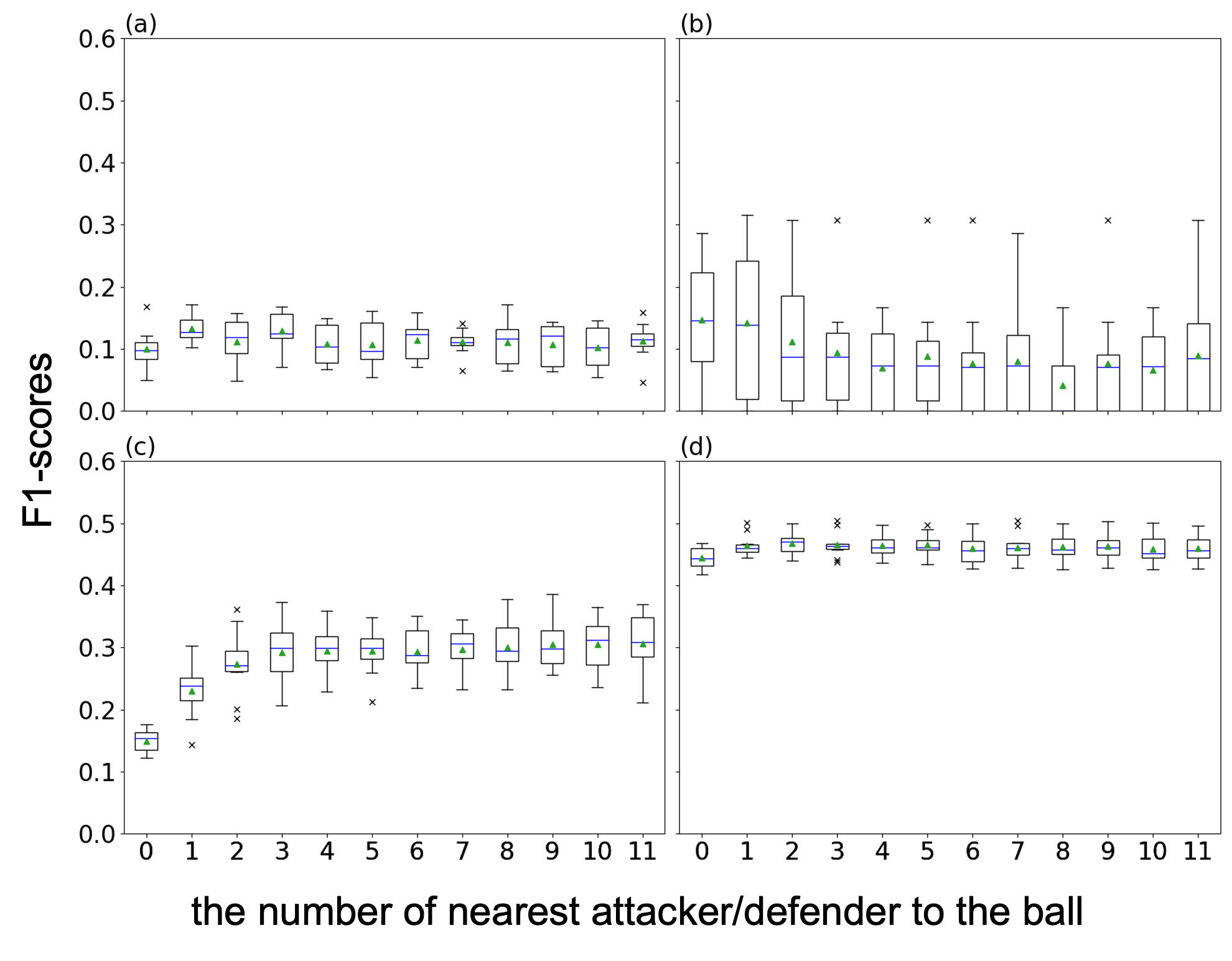}
    \caption{{\bf F1-scores of the predictions for various numbers of nearest attacker/defender to the ball ($n\_nearest$) in UEFA EURO 2020.} 
    Box plots represent F1-scores of the predictions of (a) scores, (b) concedes, (c) gains and (d) attacked for $n\_nearest$ in UEFA EURO 2020. 
    A green triangle is an average F1-score at each $n\_nearest$ and x is an outlier at a $n\_nearest$. 
    }
    \label{fig:f1scores_euro2020}
\end{figure}

To verify the GVDEP method, we first investigated the prediction performances by changes in the number of people around the ball ($n\_nearest$).
As mentioned earlier, GVDEP has four classifiers of gains, attacked, scores and concedes.
The latter two probabilities are the same as the previous work by \cite{Decroos19}.
These classifiers predict probabilities of gains ($P_{gains}$), attacked ($P_{attacked}$), scores ($P_{scores}$) and concedes ($P_{concedes}$).
In Figure \ref{fig:f1scores_euro2020}, we showed that changes in $n\_nearest$ influenced F1-scores of these predictions.
For the prediction probability of gains, 
no improvement in F1-socres was observed after $n\_nearest$ is 3 or 4.
In contrast, for the predictions of scores, concedes, and attacked, F1-scores did not increase with all players’ location information.
The results of UEFA EURO 2022 were similar and we show the results in Supplementary Figure 6.

\subsection*{Valuation of team defenses}
Next, we show examples of team defense valuations in UEFA EURO 2020 in Figure \ref{fig:valuations2020} (for UEFA EURO 2022, see Supplementary Figure 7).
For the results of the previous VDEP definition, which is formulated by $P_{gains}$ and $P_{attacked}$, see Supplementary Figure 8.
In the figures, the mean values of $\Delta P_{gains}$ and $\Delta P_{attacked}$ are defined as $gain\_value$ and $attacked\_value$ respectively, and GVDEP values are as \textit{g\_vdep\_value}.
First, we characterize and evaluate team defenses 
using \textit{gain\_value} and the average of \textit{attacked\_value} in Figure \ref{fig:valuations2020}a. 
Overall, the trade-off between the averaged gain and attacked values were confirmed ($r_{14} = -0.757, p = 0.001 < 0.05$).
That is, the teams with more gain values tended to more attacked (less value), and vice versa. 
This tendency was similar to the results of the previous work (the older definition and Japanese professional soccer league) by \cite{toda2022evaluation}.
For specific teams valuation, for example, Italy that won UEFA EURO 2020 was able to keep their opponents (Turkey, Switzerland, Wales and Austria) from penetrating into the penalty area, suggesting that these would be connected with a less number of concedes (see also Figure \ref{fig:valuations2020}b).
Also, both values of England that is one of the finalists was over each average values.
In addition, Spain and Denmark that advanced to Semi-Final were both higher \textit{attacked\_value} but lower \textit{gain\_value}, suggesting that they may keep good defense where their opposite teams was unable to attack effectively.

Second, we investigate the relationship between GVDEP values and actual concedes in Figure \ref{fig:valuations2020}b.
There was no significant correlation between them ($r_{14} = -0.265, p = 0.321$). 
Please note that the concedes were integers and had small variations, then we proposed this approach to value the defense process (not the outcome).
Thus, the difference between them is important rather than the correlation itself. 
For example, Italy and England were not conceded and were high GVDEP values, so we can guess they continued to protect themselves against concedes.

Third, it should be noted that there was a strong correlation between GVDEP and attacked values ($r_{14} = 0.993, p = 3.161\times 10^{-14}$) in Figure 3c.
This is probably because the attacked values and the absolute value of the coefficients ($|weight\_attacked| = 0.021$) were larger than those of gain values ($|weight\_gains| = 0.011$) in UEFA EURO 2020. 
Thus, we should be careful of the results interpretation in GVDEP values, which is similar to those of the attacked values. 
Specifically, the finalists tried to preserve the state where it was difficult for their opposite teams to attack effectively, so their GVDEP values are high.
Also, Spain and Denmark were at their average levels of gain and attacked values.

Lastly, for this reason, we investigated the relationship between the gain values and the concedes in Figure 3d. 
There was also no significant correlation between them ($r_{14} = 0.389, p = 0.136$).
Yet, roughly speaking, the figure shows if a team tries to bring the state where they can gain the ball in the future, they will take more risks.
For instance, the number of Italy's concedes and their \textit{gain\_value} were less as with the tendency.
In short, they did not aim to gain the ball when defending.
However, England's \textit{gain\_value} was over the mean value and their concedes were less than any other team, suggesting they tried such as tackles and interceptions and succeeded.
In addition, \textit{gain\_value} of Spain and Denmark were low but they allowed more concedes than Italy.



\section*{Discussion}
In this study, we proposed a generalized valuation method of defensive teams by score-scaling the predicted probabilities of the events.
First, we verified the existing probabilities based on the prediction performance.
Second, we quantitatively analyzed the games in UEFA EURO 2020 using the defensive evaluations of the proposed.
Finally, we discuss the limitations of the proposed methods and future perspectives.

To calculate the probabilities of ball gain and being attacked, the previous study \cite{toda2022evaluation} used data including all player's features.
However, the type of this data is not always available because this is often private or expensive.
Hence, we used open-source data including all player's location data in the video frame, and we verified the existing classifiers' performances.
This result suggests that
although features of not only the ball but the players are important to improve the classifier in ball gain, we do not necessary need all players locations. 
Our results suggest that our approach can evaluate defensive performances only with the open-source data.

Considering the team evaluations,
according to the correlation analysis, we found the trade-off between the tendencies of gaining the ball and of not being effectively attacked.
As a finalist team, England was able to maintain a good balance between both the ball gain and not being attacked at a high level, and they did not allow to concede until the analyzed Round of 16.
A champion team, Italy's $attacked\_value$ is the highest in teams that went to the knockout phase.
We also found that they did not concede except for the corner kick in the analyzed games. 
However, Belgium and Czech Republic were evaluated with lower values in our method in spite of the low number of concedes.
This may be because their keepers made efforts to prevent concedes.
Indeed, out of the four matches played up to the round of 16, Belgium and Czech Republic have kept three or two clean sheets (no concedes), respectively.
In this study, we consider the evaluation of team defense, not the contributions of the keepers, thus we may acquire such results.

\begin{figure}
    \centering
    \includegraphics[scale=0.65]{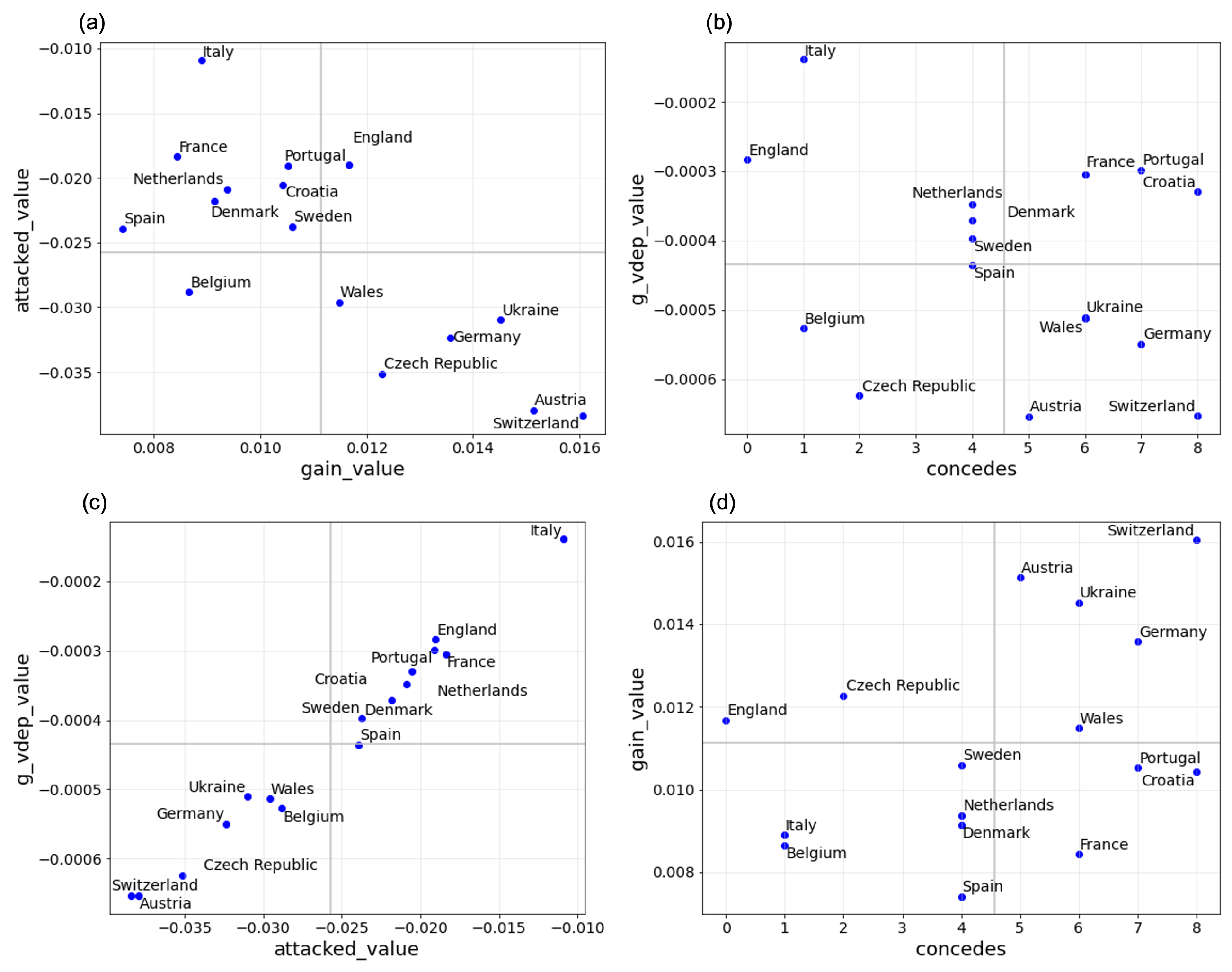}
    \caption{{\bf Defensive evaluations of teams in multiple games of UEFA EURO 2020.}
    Plots represent the relationship between (a) $gain\_value$ and $attacked\_value$, (b) concedes and GVDEP value, (c) $attacked\_value$ and GVDEP value, (d) concedes and $gain\_value$ regarding the best 16 teams of UEFA EURO 2020.
    The grey lines are the averaged values of each graph's axes. 
    For $gain\_value$, the more points plotted to the right, the more likely a team could transit to a state where they are to gain the ball.
    Also, for $attacked\_value$, the more points plotted to the right or the above, the less chance to attack effectively a team could transit to a state where they gave.
    Furthermore, for GVDEP values ($g\_vdep\_value$ in these figures), the more points plotted to the above, the better a team could perform their defense.
    For concedes, the more points plotted to the left, the better a team could perform their defense.
    }
    \label{fig:valuations2020}
\end{figure}

Finally, we introduce the limitations of this study and future perspectives.
The first is about the use of data.
Since data used in this study does not necessarily include all players' features, results about verification of classifiers do not perfectly describe the performance.
Another issue is that our formula is too affected by $attacked\_value$.
As shown in Figure \ref{fig:valuations2020}c, we found a too high positive correlation between $attacked\_value$ and GVDEP values.
It is true that not allowing the opponent to attack can be seen as reducing the probability of conceding a goal, but it is difficult to assess the defence on this basis alone.
The last is the definition of off-ball features and the modeling.
In this study, these include the x and y coordinates of positions of all players (22 players xy coordinates) and the distance and the angle of each player from the ball, sorted in the order of closest to the ball, at the analysis stage.
For future work, we can consider more specific features of the off-ball defense (e.g., a defense line) or other nonlinear modeling such as using neural networks.

\ifarxiv
\section*{Acknowledgments}
This work was supported by JSPS KAKENHI (Grant Numbers 20H04075 and 21H05300), JST START University Ecosystem Promotion Type (Grant Number JPMJST2183), and JST Presto (Grant Number JPMJPR20CA).

\input{reference.bbl}
\else
\bibliographystyle{apa}
\bibliography{reference}
\fi

\newpage

\input{Supplementary_materials} 
\end{document}

%% file: Supplementary_materials.tex
\newif\ifarxiv
\arxivfalse 

\ifarxiv
\renewcommand{\thesection}{\Alph{section}}
\setcounter{section}{0}
\setcounter{figure}{0}
\setcounter{table}{0}
\else



\fi

\newpage


\begin{figure}[h]  
    \begin{center}
    \vspace{30pt}
    \includegraphics[scale=0.7]{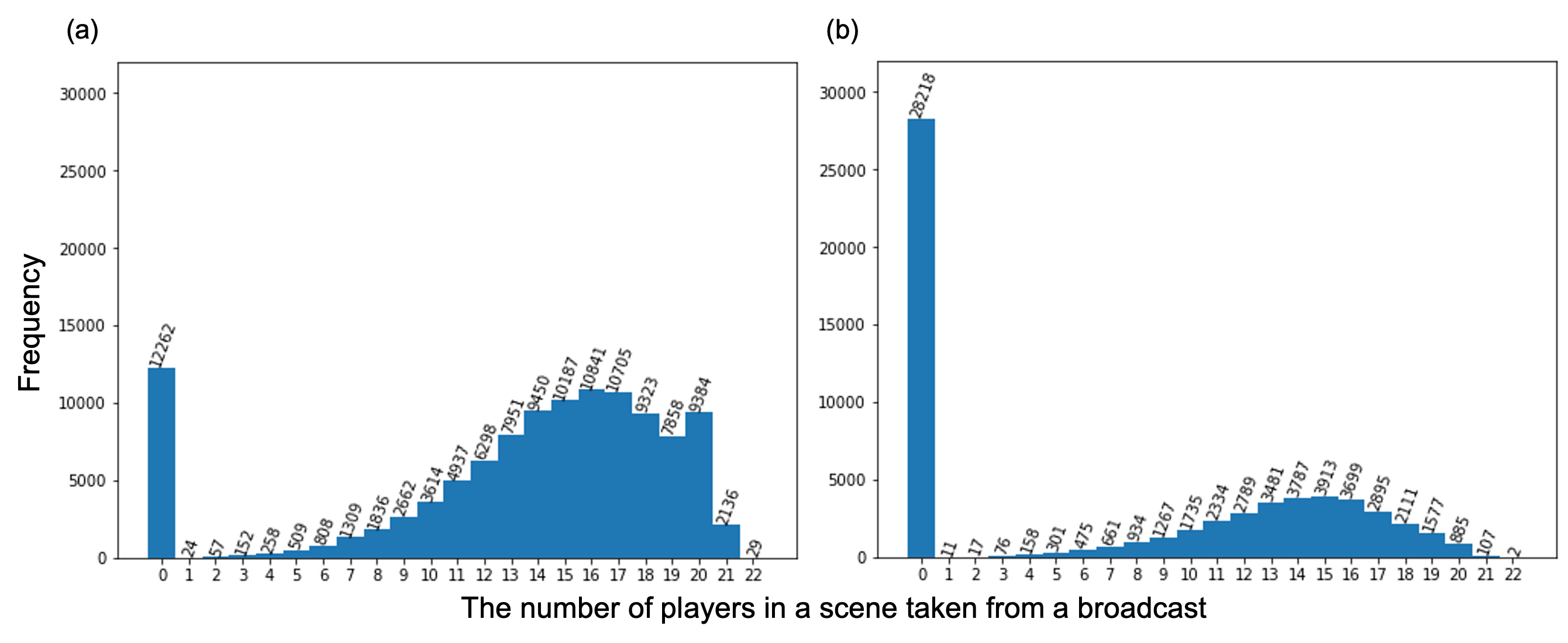}	
    \caption{
    {\bf The number of players in broadcast videos of Euro 2020 and Euro 2022.}
    This figure shows how many players there were in broadcast videos of (a) Euro 2020 and (b) Euro 2022. 
    There were no players in some scenes. 
    In this study, to calculate the probabilities of the predictions, we removed the data (12,262 in Euro 2020 and 28,218 in Euro 2022) that did not have any player in a scene.
    } 
    \label{fig:environment}
    \end{center}
\end{figure}

\newpage

\begin{figure}[h]  
    \centering
    \includegraphics[scale=0.8]{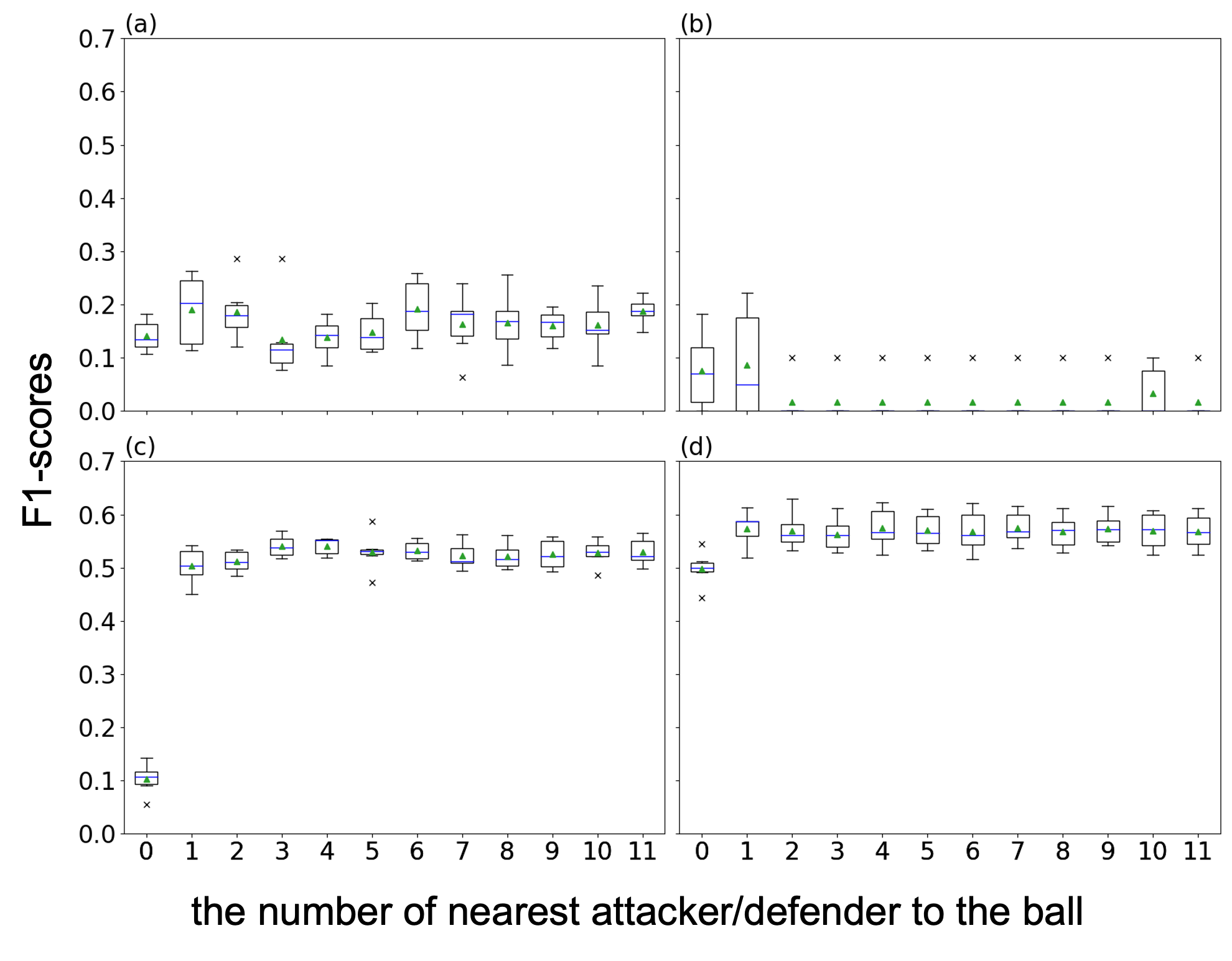}
    \caption{
    {\bf F1-scores of the predictions for n\_nearest in Euro 2022.} 
    Box plots represent F1-scores of the predictions of (a) scores, (b) concedes, (c) gains and (d) attacked for n\_nearest in Euro 2022.
    As with Figure \ref{fig:f1scores_euro2020}, a green triangle means an average of F1-score at each n\_nearest and x is an outlier at a n\_nearest.
    In particular, (c), F1-scores stop to increase when n\_nearest is 3 or 4 like Euro 2020.
    }
    \label{fig:f1scores_euro2022}
\end{figure}

\newpage

\begin{figure}
    \centering
    \includegraphics[scale=0.65]{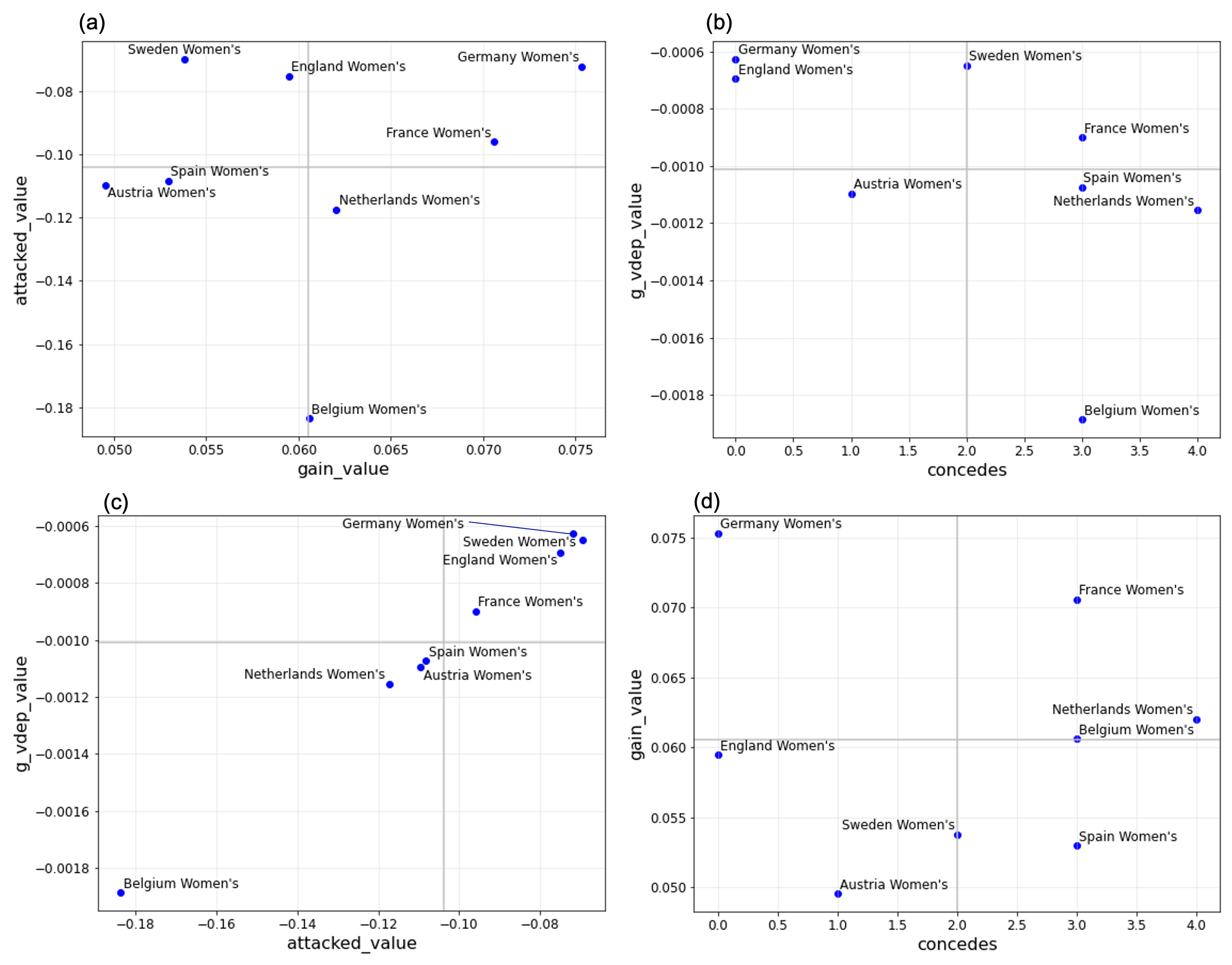}
    \caption{
    {\bf Defensive evaluations of teams in multiple games of Euro 2022.}
    Plots represent the relationship between (a) $gain\_value$ and $attacked\_value$ ($r_{14} = 0.156, p = 0.712$), (b) concedes and GVDEP values ($r_{14} = -0.557, p = 0.151$), (c) $attacked\_value$ and GVDEP values ($r_{14} = 0.999, p = 3.39\times 10^{-9}$), (d) concedes and $gain\_value$ ($r_{14} = -0.116, p = 0.784$) regarding last 8 teams of Euro 2022.
    The details about each axis were already explained in Figure \ref{fig:valuations2020}.
    As with England that was one of the finalists of Euro 2020, Germany Women's that was one of the finalists of Euro 2022 took high values for $gain\_value$ and $attacked\_value$.
    Also, similar to Italy that was another finalist of Euro 2020, another finalist of Euro 2022, England Women's, was able to keep their opponents from penetrating into the penalty area.
    These would be connected with a less number of concedes (see also Figure \ref{fig:valuations2022}b).
    }
    \label{fig:valuations2022}
\end{figure}

\newpage

\begin{figure}[h]  
    \begin{center}
    \vspace{30pt}
    \includegraphics[scale=0.65]{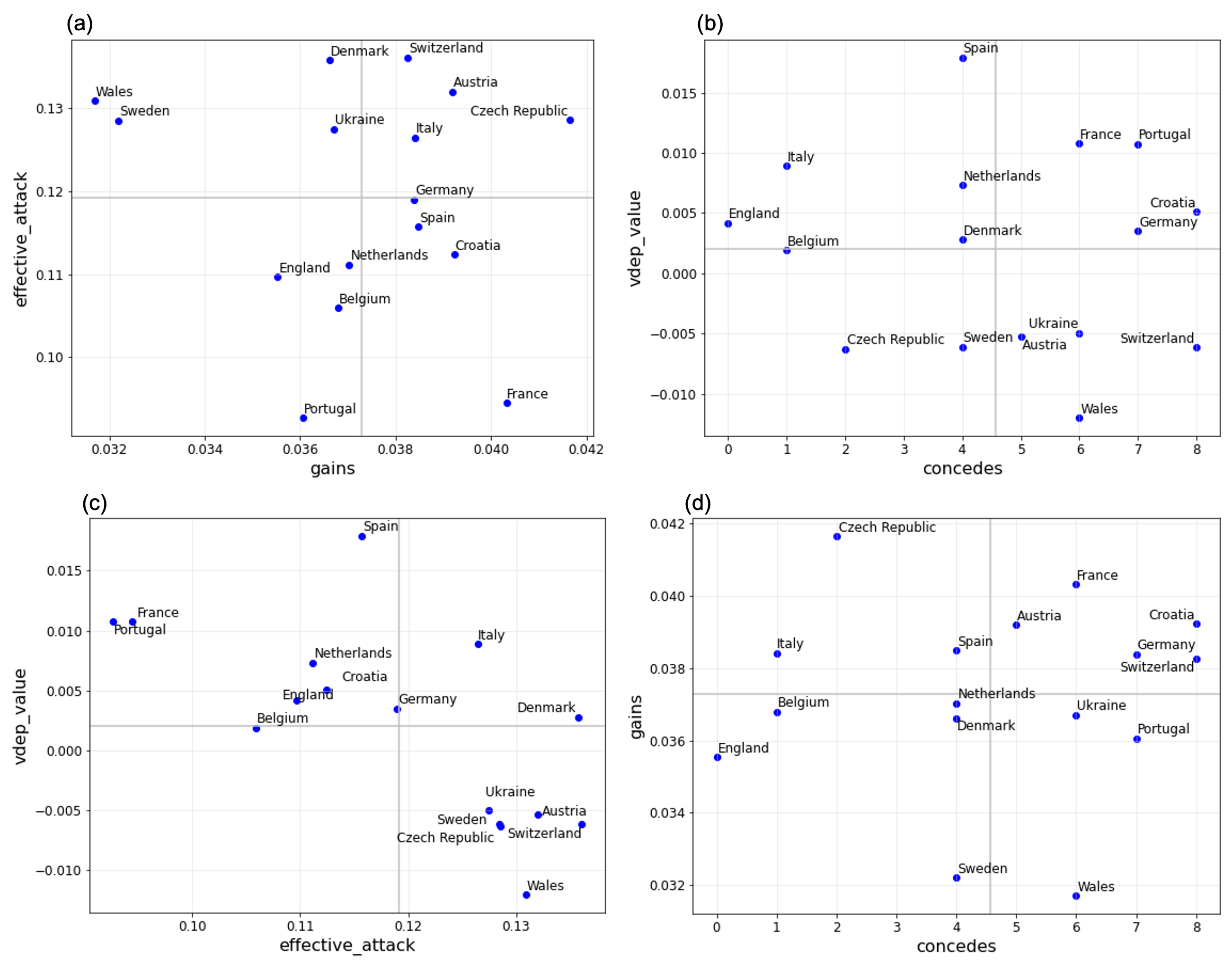}	
    \caption{
    {\bf Defensive evaluations of teams in multiple games of Euro 2020 in the definition of \cite{toda2022evaluation}.}
    Plots representing the relationship between (a) $P_{gains}$ and $P_{attacked}$ ($r_{14} = -0.135, p = 0.617$), (b) concedes and VDEP values ($r_{14} = -0.110, p = 0.686$), (c) $P_{attacked}$ and VDEP values ($r_{14} = -0.669, p = 0.00463$), (d) concedes and $P_{gains}$ ($r_{14} = 0.0333, p = 0.903$) regarding last 16 teams of Euro 2020.
    he grey lines are the averaged values of each graph's axes. 
    For $P_{gains}$, the more points plotted to the right, the more likely a team could be to gain the ball.
    Yet, for $P_{attacked}$, the more points plotted to the left or the below, the less likely an opposite team could be to attack effectively.
    Otherwise, the axes are the same meanings as Figure \ref{fig:valuations2020}.
    VDEP value that was proposed by \cite{toda2022evaluation} is calculated as follows: $P_{gains} - C \times P_{attacked}$.
    Since $C$ value is based on the number of occurrences of ball gains and being attacked,  this varies with the data ($C = 0.313$ in this study).
    } 
    \label{fig:res_testdata}
    \end{center}
\end{figure}


%% file: main.bbl
\begin{thebibliography}{}

\bibitem[\protect\astroncite{Chen and Guestrin}{2016}]{Chen16}
Chen, T. and Guestrin, C. (2016).
\newblock Xgboost: A scalable tree boosting system.
\newblock In {\em Proceedings of the ACM SIGKDD International Conference on
  Knowledge Discovery \& Data Mining}, pages 785--794.

\bibitem[\protect\astroncite{Cioppa et~al.}{2022}]{Cioppa_2022_CVPR}
Cioppa, A., Giancola, S., Deli\`ege, A., Kang, L., Zhou, X., Cheng, Z., Ghanem,
  B., and Van~Droogenbroeck, M. (2022).
\newblock Soccernet-tracking: Multiple object tracking dataset and benchmark in
  soccer videos.
\newblock In {\em Proceedings of the IEEE/CVF Conference on Computer Vision and
  Pattern Recognition (CVPR) Workshops}, pages 3491--3502.

\bibitem[\protect\astroncite{Decroos et~al.}{2019}]{Decroos19}
Decroos, T., Bransen, L., Van~Haaren, J., and Davis, J. (2019).
\newblock Actions speak louder than goals: Valuing player actions in soccer.
\newblock In {\em Proceedings of the 25th ACM SIGKDD International Conference
  on Knowledge Discovery \& Data Mining}, pages 1851--1861.

\bibitem[\protect\astroncite{Fernandez and Bornn}{2018}]{Fernandez18}
Fernandez, J. and Bornn, L. (2018).
\newblock Wide open spaces: A statistical technique for measuring space
  creation in professional soccer.
\newblock In {\em Proceedings of the 12th MIT Sloan Sports Analytics
  Conference}.

\bibitem[\protect\astroncite{Friedman}{2001}]{friedman2001greedy}
Friedman, J.~H. (2001).
\newblock Greedy function approximation: a gradient boosting machine.
\newblock {\em Annals of statistics}, pages 1189--1232.

\bibitem[\protect\astroncite{Fujii}{2021}]{fujii2021data}
Fujii, K. (2021).
\newblock Data-driven analysis for understanding team sports behaviors.
\newblock {\em Journal of Robotics and Mechatronics}, 33(3):505--514.

\bibitem[\protect\astroncite{Gregory et~al.}{2022}]{gregory2022influence}
Gregory, S., Robertson, S., Aughey, R., and Duthie, G. (2022).
\newblock The influence of tactical and match context on player movement in
  football.
\newblock {\em Journal of Sports Sciences}, 40(9):1063--1077.

\bibitem[\protect\astroncite{Liu et~al.}{2020}]{Liu2020}
Liu, G., Luo, Y., Schulte, O., and Kharrat, T. (2020).
\newblock Deep soccer analytics: learning an action-value function for
  evaluating soccer players.
\newblock {\em Data Mining and Knowledge Discovery}, 34(5):1531--1559.

\bibitem[\protect\astroncite{Pappalardo et~al.}{2019}]{pappalardo2019playerank}
Pappalardo, L., Cintia, P., Ferragina, P., Massucco, E., Pedreschi, D., and
  Giannotti, F. (2019).
\newblock Playerank: data-driven performance evaluation and player ranking in
  soccer via a machine learning approach.
\newblock {\em ACM Transactions on Intelligent Systems and Technology (TIST)},
  10(5):1--27.

\bibitem[\protect\astroncite{Piersma}{2020}]{Piersma20}
Piersma, J. (2020).
\newblock Valuing defensive performances of football players.
\newblock {\em Master Thesis in Erasmus School of Economics}.

\bibitem[\protect\astroncite{Rahimian et~al.}{2022}]{rahimian2022lets}
Rahimian, P., Gomes, D., Berkovics, F., and Toka, L. (2022).
\newblock Let's penetrate the defense: A machine learning model for prediction
  and valuation of penetrative passes.
\newblock In {\em International Workshop on Machine Learning and Data Mining
  for Sports Analytics}. Springer.

\bibitem[\protect\astroncite{Robberechts}{2019}]{Robberechts19}
Robberechts, P. (2019).
\newblock Valuing the art of pressing.
\newblock In {\em StatsBomb Innovation in Football Conference}.

\bibitem[\protect\astroncite{Rudd}{2011}]{rudd2011framework}
Rudd, S. (2011).
\newblock A framework for tactical analysis and individual offensive production
  assessment in soccer using markov chains.
\newblock In {\em New England symposium on statistics in sports}.

\bibitem[\protect\astroncite{Scott et~al.}{2022}]{scott2022soccertrack}
Scott, A., Uchida, I., Onishi, M., Kameda, Y., Fukui, K., and Fujii, K. (2022).
\newblock Soccertrack: A dataset and tracking algorithm for soccer with
  fish-eye and drone videos.
\newblock In {\em Proceedings of the IEEE/CVF Conference on Computer Vision and
  Pattern Recognition}, pages 3569--3579.

\bibitem[\protect\astroncite{Spearman}{2018}]{spearman2018beyond}
Spearman, W. (2018).
\newblock Beyond expected goals.
\newblock In {\em Proceedings of the 12th MIT Sloan Sports Analytics
  Conference}, pages 1--17.

\bibitem[\protect\astroncite{Teranishi et~al.}{2022}]{teranishi2022evaluation}
Teranishi, M., Tsutsui, K., Takeda, K., and Fujii, K. (2022).
\newblock Evaluation of creating scoring opportunities for teammates in soccer
  via trajectory prediction.
\newblock In {\em International Workshop on Machine Learning and Data Mining
  for Sports Analytics}. Springer.

\bibitem[\protect\astroncite{Toda et~al.}{2022}]{toda2022evaluation}
Toda, K., Teranishi, M., Kushiro, K., and Fujii, K. (2022).
\newblock Evaluation of soccer team defense based on prediction models of ball
  recovery and being attacked: A pilot study.
\newblock {\em Plos one}, 17(1):e0263051.

\bibitem[\protect\astroncite{Van~Roy et~al.}{2020}]{vanroy2020valuing}
Van~Roy, M., Robberechts, P., Decroos, T., and Davis, J. (2020).
\newblock Valuing on-the-ball actions in soccer: a critical comparison of xt
  and vaep.
\newblock In {\em Proceedings of the AAAI-20 Workshop on Artifical Intelligence
  in Team Sports}. AI in Team Sports Organising Committee.

\end{thebibliography}
